\documentclass[11pt,a4paper]{article}
\usepackage[hyperref]{emnlp2020}
\usepackage{times}
\usepackage{latexsym}

\usepackage{microtype}
\usepackage{graphicx}
\usepackage{amsmath}
\usepackage{amsfonts}
\usepackage{booktabs}
\usepackage{algorithm}
\usepackage{algorithmic}
\usepackage{bm}
\usepackage{amssymb}
\usepackage{natbib}

\aclfinalcopy 
\def\aclpaperid{2975} 

\title{Counterfactual Off-Policy Training for Neural Dialogue Generation}

\author{Qingfu Zhu$^\sharp$, Weinan Zhang$^{\sharp}$\thanks{\ \ Corresponding author.}, Ting Liu$^{\sharp }$,  William Yang Wang$^{\flat}$ \\
  $^\sharp$Harbin Institute of Technology, Harbin, China \\
  $^\flat$University of California, Santa Barbara, USA\\
  {\tt \{qfzhu, wnzhang, tliu\}@ir.hit.edu.cn} \\
  {\tt william@cs.ucsb.edu} \\
  }
  
\date{}

\begin{document}
    \maketitle
    \begin{abstract}
Open-domain dialogue generation suffers from the data insufficiency problem due to the vast size of potential responses.
In this paper, we propose to explore potential responses by counterfactual reasoning.
Given an observed response, the counterfactual reasoning model automatically infers the outcome of an alternative policy that could have been taken.
The resulting counterfactual response synthesized in hindsight is of higher quality than the response synthesized from scratch. 
Training on the counterfactual responses under the adversarial learning framework helps to explore the high-reward area of the potential response space.
An empirical study on the DailyDialog dataset shows that our approach significantly outperforms the HRED model as well as the conventional adversarial learning approaches.
\end{abstract}

    \section{Introduction}
    Open-domain dialogue generation~\cite{shang-etal-2015-neural,vinyals2015seq,sordoni-etal-2015-neural} intends to produce coherent responses given dialogue history.
    Nevertheless, it suffers from data insufficiency problem as there may exist many potential responses for a given dialogue history~\cite{li2016mmi}.
    An ideal way of exploring the potential responses is to train the model by chatting with real users, which is usually time-consuming and labor-intensive in practice.
    Although replacing a real user with a user simulator could address the issue,
    the simulator only roughly approximates real user statistics, and its development process is costly~\cite{su-etal-2016-line}.
   
    \begin{figure}[!t]
        \centering
        \includegraphics[width=220pt]{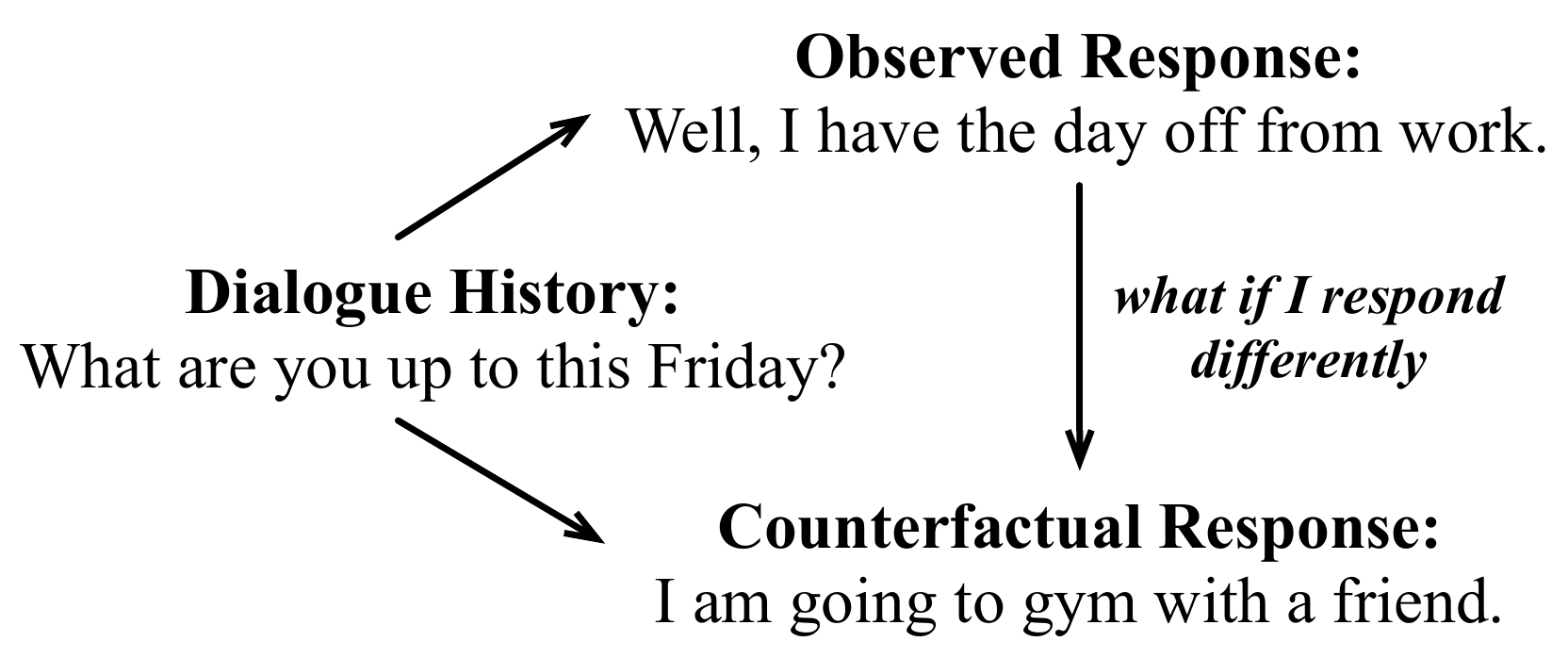}
        \caption{\label{ratio} 
            An example of a counterfactual response, which is a potential response inferred in hindsight from given observed response.
        }
    \end{figure}
    
    In contrast, humans could independently reason potential responses based on past experiences from the true environment.
    Having observed a response, one might naturally ask himself or herself: {``\it What would happen if I respond differently, while everything else in the environment remains the same.''}
    Answering the question will result in a potential response (as an example in Figure~\ref{ratio}), and it is beneficial for improving future decision making~\cite{roese1997counterfactual}.
    The potential response inferred in hindsight is called a {\it counterfactual} response, where the concept ``counterfactual'' describes the posterior process of reasoning the outcome of alternative actions (i.e., a different responding policy) that could have been taken while keeping everything else unchanged~\cite{buesing2018woulda}.

    Motivated by this, we propose a counterfactual off-policy training (COPT) approach to explore potential responses.
    Building upon the adversarial learning framework, COPT casts a dialogue generator as a structural causal model (SCM), which describes a generation process with two ingredients: {\it scenarios} and {\it causal mechanisms}~\cite{wright1920relative,buesing2018woulda}.
    The scenario is a random noise variable that captures all unobserved yet relevant aspects of the environment, i.e., user profiles.
    The causal mechanism is a deterministic function that takes a scenario and dialogue history as input and outputs a response.
    In this way, reasoning a counterfactual response in an observed response's environment can be achieved by feeding the scenario of the observed response into the causal mechanism.
    After generating the counterfactual response, the generator will receive a reward from a discriminator and optimize itself accordingly.

    Intuitively, a counterfactual response is synthesized by grounding the model in the scenario where an observed response occurs, rather than the scenario sampled from scratch as standard adversarial learning-based approaches.
    This improves the quality of the synthesized responses and subsequently benefits the model that learns from the synthesis.
    To verify the effectiveness of our approach, we conduct experiments on the public available DailyDialog dataset~\cite{li2017dailydialog}.
    Experimental results show that our approach significantly outperforms previous adversarial learning-based approaches in both automatic and human evaluations.
    The contributions of this paper are summarized as follows:
    \begin{itemize}
        \item We connect the concept of counterfactual reasoning with the dialogue generation by casting the dialogue generation model as a structural causal model.
        \item Our counterfactual response is of higher quality than the response synthesized from scratch in standard adversarial learning-based dialogue generation model.
        \item Our approach is model-agnostic and can be applied to any adversarial learning-based dialogue generation model.
    \end{itemize}
    
    \section{Related Work}
    \paragraph{Dialogue Generation}
    Data-driven dialogue systems can be roughly divided into two categories: retrieval-based~\cite{leuski2009building,ji2014information,yan2016learning}  and generation-based~\cite{shang2015neural,sordoni2015seq,vinyals2015seq}.
    Responses of retrieval-based methods come from a fixed candidate response set and thus are incapable of being customized.
    The generation-based methods can create new responses, but the vanilla sequence to sequence model tends to produce generic responses~\cite{li2016mmi}.
    
    One way to address the generic response problem is by introducing external knowledge, such as keywords~\cite{mou2016seq2bf,zhu2019order}, topics~\cite{xing2017topic}, persona information~\cite{zhang2019neural,songexploiting}, and retrieved candidate responses~\cite{song2018ensemble,wu2019response,zhu-etal-2019-retrieval}.
    Another way is to optimize the architecture of networks.
    There are two architectures widely employed in this research line: the variational auto-encoder~\cite{bowman2015generating,zhao-etal-2017-learning} and the generative adversarial network~\cite{goodfellow2014generative,li2017adversarial,zhang2018generating,xu2018diversity,tuan2019improving}.
    Our approach falls into the latter category.
    The differences between our approach and other adversarial learning-based approaches are as follows.
    First, we cast the dialogue generation model as an SCM to explore potential responses in the environment where observed responses occur.
    Second, we learn on counterfactual responses that inferred from the SCM.
    Third, a pre-trained behavior policy is involved during the generation process, making our approach an off-policy algorithm and benefits the exploration of potential responses.

    \paragraph{Counterfactual Reasoning}
    The counterfactual reasoning is a concept derived from psychology.
    It describes the human capacity to learn from experience by reasoning the outcome of an alternative action that could have been taken~\cite{pearl2018book}.
    Combined with the SCM, counterfactual reasoning improves the performance of policy evaluation in reinforcement learning~\cite{buesing2018woulda,oberst2019counterfactual}.
    In the area of NLP, counterfactual reasoning in previous work is mainly used for data augmentation~\cite{qin-etal-2019-counterfactual,fu2019counterfactual,kaushik2019learning}, which rewrites the original data given a counterfactual label or condition.
    In this paper, we connect the concept of counterfactual reasoning with the dialogue generation and are the first to cast a generation model as an SCM under the adversarial learning framework.
    
    \section{Method}
    We cast a dialogue generation model as an SCM to explore potential responses by counterfactual reasoning during the training process.
    We will first review the concept of the SCM (Sec.~\ref{sec_scm}), and then introduce our COPT approach (Sec.~\ref{sec_copt}).
    
    \subsection{Notation}
    We use capital letters for random variables (e.g., $V$), lowercase letters for instances of random variables (e.g., $v$), and bold letters for vectors (e.g., $\bm{V}=\{V_1,..., V_N\}$).
    During the training process, we denote the response generated by COPT as {\it counterfactual response}.
    In contrast, the response of standard adversarial learning-based dialogue generation (i.e., REGS~\citealp{li2017adversarial}) is denoted as {\it standard response}.
    
    \begin{figure}[!t]
        \centering
        \includegraphics[width=200pt]{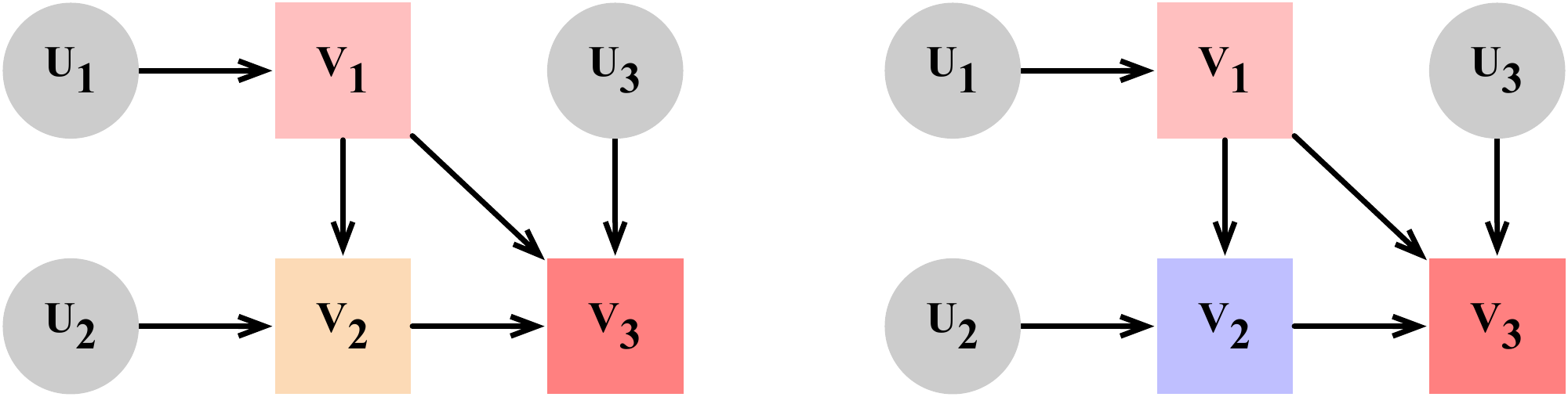}
        \caption{\label{SCM} 
            An example of an SCM and an intervention. \textbf{Left:} An SCM with random variables $\bm{V}$, scenarios $\bm{U}$, and causal mechanisms $\bm{F}$ represented by colored squares.
            \textbf{Right:} A new SCM after taking an intervention on the left SCM. The original causal mechanism $f_2(\bm{V}_1,\bm{U}_2)$ (denoted by the orange square) is replaced by $f_2^T(\bm{V}_1,\bm{U}_2)$ (denoted by the purple square).
        }
    \end{figure}
    
    \subsection{Background: Structural Causal Model} \label{sec_scm}
    A structural causal model over random variables $\bm{V}=\{\bm{V}_1, ..., \bm{V}_N\}$ consists of
    independent noise random variables $\bm{U}=\{\bm{U}_1, ..., \bm{U}_N\}$ with distribution $P_{\bm{U}}$ and
    deterministic functions $\bm{F}=\{f_1, ..., f_N\}$ such that $\bm{V}_i = f_i(\bm{PA}_i, \bm{U}_i)$,
    where $\bm{PA}_i \subset \bm{V}$ are the parents of $\bm{V}_i$ in a given DAG~\cite{buesing2018woulda}.
    $\bm{U}$ is called {\it scenarios}, and $\bm{F}$ is called {\it causal mechanisms}.
    Figure~\ref{SCM} (Left) shows an example of an SCM.
    Each random variable $\bm{V}_i$ is determined by its parents in $\bm{V}$, $\bm{U}_i$, and $f_i$, e.g., $\bm{V}_2 = f_2(\bm{V}_1, \bm{U}_2)$.
    
    During the training process, we cast a dialogue generation model as an SCM over two random variables: dialogue history $\bm{X}$ and response $\bm{Y}$.
    This is achieved by converting the conditional distribution $P(\bm{Y}|\bm{X})$ into a deterministic function $\bm{Y}=f_{\pi}(\bm{X}, \bm{U})$ (for more details see Sec.~\ref{sec_copt}).
    The scenario $\bm{U}$ is a random noise variable that captures all unobserved yet relevant properties, like user profiles.
    The causal mechanism is denoted as $f_{\pi}$ to highlight the role of the policy (parameters)  $\pi$ of the model.
    The dialogue generation SCM makes it possible to sample counterfactual responses in the scenario where observed responses occur.
    This improves the quality of synthesized responses and subsequently helps the model to explore the high-reward area of the potential response space in the training process.

    \begin{figure*}[!t]
        \centering
        \includegraphics[width=380pt]{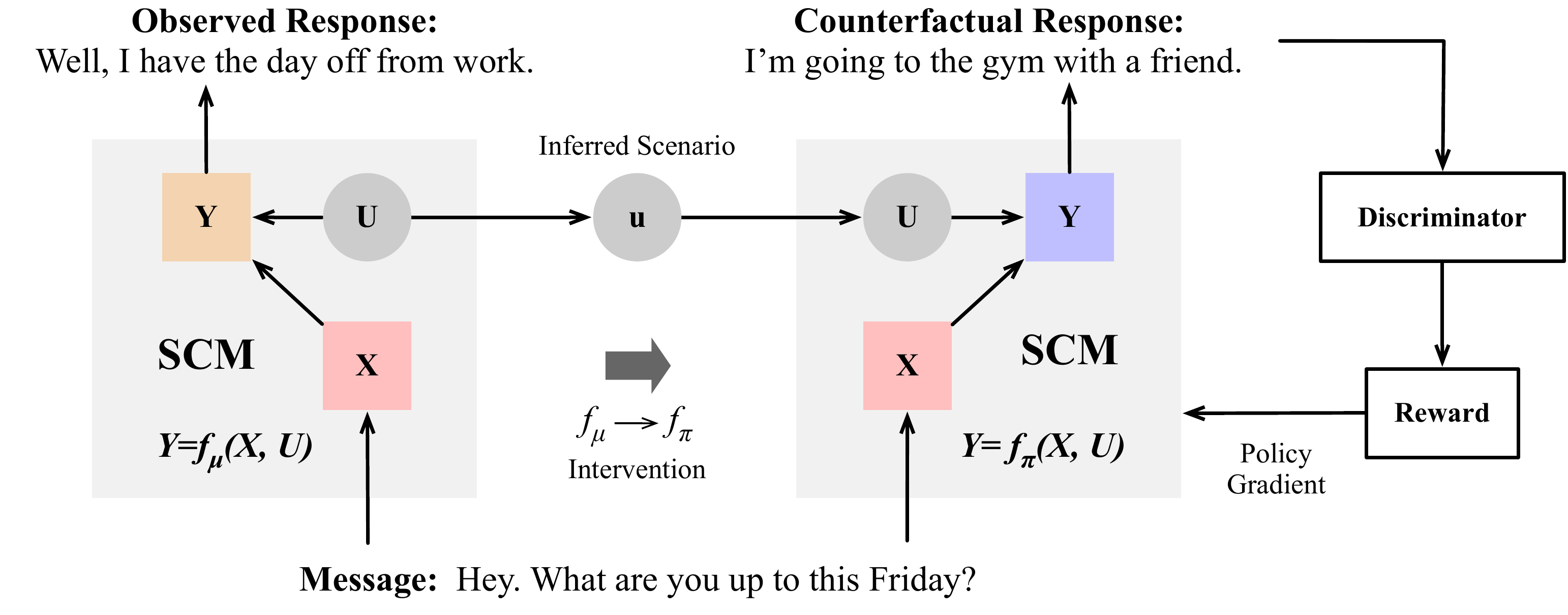}
        \caption{\label{model} 
            The architecture of our COPT approach.
            $\pi$ is the target policy that we aim to learn.
            $\mu$ is the behavior policy that generates observed responses.
            First, we infer the
            scenario $\bm{u}$ where the observed response occurs.
            Then we update the policy from $\mu$ to $\pi$, which can be seen as an intervention on the left SCM and results in the right SCM.
            Then, the counterfactual response is reasoned in the inferred scenario $\bm{u}$ by the causal mechanism $\bm{Y}=f_\pi(\bm{X}, \bm{U})$.
        }
    \end{figure*}
    
    \paragraph{Intervention in SCM} Given an SCM, an intervention $T$ is defined as the replacement of some causal mechanisms.
    Figure~\ref{SCM} shows an example of intervention.
    The original causal mechanism $ f_2(\bm{V}_1,\bm{U}_2)$ in the left SCM is replaced with $f_2^T(\bm{V}_1,\bm{U}_2)$, resulting in a new SCM in the right.
    Accordingly, intervention in our dialogue generation SCM denotes the update of the policy.
    For instance, the update from the behavior policy $\mu$ that generates observed responses to the target policy $\pi$ that we aim to learn is the intervention of replacing $f_{\mu}(\bm{X}, \bm{U})$ with $f_{\pi}(\bm{X}, \bm{U})$.
    
    \paragraph{Counterfactual Reasoning in SCM}
    Given an SCM and observed a variable $\bm{V}_i=\bm{v}_i$, counterfactual reasoning answers the question:
    {\it ``What the variable $\bm{V}_i$ would have been if I take an intervention $T$ while remaining everything else unchanged''}.
    In this way, generating a counterfactual response can be seen as querying:
    {\it ``Having observed a response $\bm{Y}=\bm{y}$, what the response $\bm{Y}$ would have been if I take an intervention by following the target policy $\pi$, rather than the behavior policy $\mu$ that generates the observed responses''}.
    
    Typically, counterfactual reasoning answers the question by the following steps (as Figure~\ref{model}):
    \begin{itemize}
        \item Observed $\bm{Y}=\bm{y}$ when $\bm{X}=\bm{x}$, infer the scenario $\bm{u}$ in hindsight from $\bm{Y}=f_{\mu}(\bm{X},\bm{U})$. 
        \item Take an intervention by replacing the causal mechanism $f_{\mu}(\bm{X}, \bm{U})$ with $f_{\pi}(\bm{X}, \bm{U})$.
        \item Reason a counterfactual response $\bm{\hat{y}} = f_{\pi}(\bm{x}, \bm{u})$ by the resulting new SCM. 
    \end{itemize}
    In the following sections, we denote an observed response from the training set as $\bm{Y}$ and a model-generated response as $\bm{\hat{Y}}$.
    
    \subsection{Counterfactual Off-Policy Training} \label{sec_copt}
    Our COPT approach is model-agnostic and can be applied to any adversarial learning-based dialogue generation model.
    Without loss of generality, we take the combination of COPT and the reward for every generation step (REGS) model~\cite{li2017adversarial} as an example in this section.
    It consists of two main components: a generator $G$ and a discriminator $D$.
    
    \paragraph{Generator}
    The generator $G$ is a sequence to sequence (Seq2Seq) model~\cite{sutskever2014seq} equipped with the attention mechanism~\cite{bahdanau2014align,luong2015effective}.
    During the encoding process, $G$ reads the dialogue history into hidden states using an encoder LSTM~\cite{hochreiter1997long}:
    \begin{equation} \label{enc}
    \bm{H}_i = \text{LSTM}(X_i, \bm{H}_{i-1}),
    \end{equation}
    where $X_i$ is the $i$-th word of the dialogue history,
    and $\bm{H}_i$ denotes the corresponding hidden state.
    At the $j$-th decoding time step,
    the hidden states are summarized into a context vector $\bm{C}_j$ by the attention mechanism.
    Subsequently, $G$ predicts the distribution of the next word over the vocabulary by a decoder LSTM:
    \begin{align} \label{dec}
    &\bm{S}_j = \text{LSTM}([\,\bm{e}(\hat{Y}_{j-1}), \bm{C}_j\,], \bm{S}_{j-1}),\\
    &\bm{P}_j^{\pi}(\hat{Y}_j|\bm{X}, \bm{\hat{Y}}_{1:j-1}) = \text{softmax}(\bm{S}_{j} \cdot \bm{O}), \label{sf}
    \end{align}
    where the bracket [$\cdot$,$\cdot$] denotes concatenation, and $\bm{e}(\cdot)$ denotes the embedding of a word.
    $\bm{S}_j$ is the $j$-th hidden state of the decoder LSTM.
    $\hat{Y}_{j-1}$ is the word generated in the previous time step.
    $\bm{O}$ is the output matrix.
    We use the superscript in $\bm{P}_j^{\pi}$ to highlight the role of the policy (parameters of $G$).
    
    Adversarial learning-based dialogue generation model is optimized according to the reward of responses sampled from $\bm{P}_j^{\pi}(\hat{Y}_j|\bm{X}, \bm{\hat{Y}}_{1:j-1}) \in \mathbb{R}^{|V|}$ (abbreviated as $\bm{P}_j^\pi$ in the following), where $|V|$ is the vocabulary size.
    Using the Gumbel-Max Trick~\cite{luce2012individual},
    the sampling process can be achieved by:
    \begin{align}
    \hat{Y}_j=&{{\arg\max}_{\hat{Y}_j}}(\log \bm{P}_j^{\pi}+\bm{U}_j),\label{ctf}
    \end{align}
    where the element of $\bm{U}_j$ follows the standard Gumbel distribution.
    In this way, the generator turns into a Gumbel-Max SCM~\cite{oberst2019counterfactual}, whose scenarios and causal mechanisms are represented by $\bm{U}_j$ and Equation~\ref{ctf}, respectively.
    
    From the perspective of the SCM,
    each response is generated in a scenario.
    For instance, a standard response is produced in a scenario sampled from scratch.
    In contrast, the scenario for a counterfactual response is inferred from an observed response $\bm{y}=\{y_j|\,{y}_j={\arg\max}_{y_j}(\log \bm{p}_j^{*}+\bm{u}_j)\}$, where * is the user's policy that generates the observed response in the true environment.
    However, the user's policy is not available in practice, which hinders the posterior inference of the scenario.
    To this end, we introduce a behavior policy $\mu$ instead and learn it by minimizing the MLE loss on observed responses.
    In this way, an observed response can be seen as generating in a scenario $\bm{u}_j$ while following the policy $\mu$: $y_j={\arg\max}_{y_j}(\log \bm{p}_j^{\mu} + \bm{u}_j)$.
    
    According to \newcite{oberst2019counterfactual}, there are two ways to infer the scenario $\bm{u}_j$ in hindsight from $y_j={\arg\max}_{y_j}(\log \bm{p}_j^{\mu} + \bm{u}_j)$ given $y_j$ and $\mu$.
    One way is the rejection sampling, which samples $\bm{u}_j$ from the standard Gumbel distribution and rejects those where $y_j \ne {\arg\max}_{y_j}(\log \bm{p}_j^{\mu} + \bm{u}_j)$.
    The other way of the posterior inference makes use of the properties of the shifted Gumbel $\bm{g}=\log \bm{p}_j^{\mu} + \bm{u}_j$:
    the maximum of $\bm{g}$ follows the standard Gumbel distribution and is independent with the argmax of $\bm{g}$~\cite{maddison2014sampling}.
    Therefore, $\bm{g}$ can be obtained by first sampling a maximum and then sampling the remaining elements truncated at the maximum.
    And $\bm{u}_j$ is subsequently computed by subtracting $
    \log\bm{p}_j^\mu$ from $\bm{g}$.
    We employ the second method to infer the scenario in COPT because it is more time-efficient than rejection sampling\footnote{In our experiments, using the second sampling method takes 0.79 seconds training on a batch when the batch size is 64. In contrast, the rejection sampling takes roughly 1.45 hours, making it hard to be used in practice.}.
    
    Given the scenario inferred from the observed response, COPT reasons the counterfactual response by feeding the dialogue history and the scenario into the SCM (Equation~\ref{ctf}).
    Then the discriminator evaluates the counterfactual response and returns a reward to the generator.
    Note that the counterfactual response and the SCM are utilized for the training process.
    During the inference process, responses are generated in the same way as the standard adversarial learning-based dialogue generation (beam search or sampling from $\bm{P}_j^{\pi}$, we use the former in our approach) because the observed response is not available.
    
    \paragraph{Discriminator} \label{disc}
    The discriminator $D$ provides a reward for each generation step.
    It takes as input the dialogue history $\bm{X}$, the word ${\tilde{Y}}_j$ produced in the current generation step, and the prefix $\bm{\tilde{Y}}_{1:j-1}$ in previous steps, where $\tilde{\bm{Y}} \in \{\bm{Y},\bm{\hat{Y}}\}$ can be either an observed response or a model-generated response.
    The output reward $D(\tilde{Y}_j|\bm{X}, \bm{\tilde{Y}}_{1:j-1})$ is the probability that $\tilde{Y}_j$ is human-generated.
    Concretely, $D$ first reads $\bm{X}$ and $\bm{\tilde{Y}}_{1:j}$ with an encoder-decoder model.
    Then, it computes the reward by a Multi-Layer Perceptron (MLP), which takes as input the last hidden state of the decoder.
    
    \paragraph{Adversarial Learning} \label{sec_adv}
    We train $G$ and $D$ under the adversarial learning framework, where $G$ tries to fool $D$ by generating human-like responses while $D$ aims to distinguish between model-generated and human-generated (the observed) responses.
    Since a response is a sequence of discrete tokens, we pass by the gradient of $D$ to $G$ using the policy gradient algorithm.
    In this way, $G$ converts into an agent whose partially generated response and parameters define a state and a policy, respectively.
    At each generation step, the agent takes an action by producing a word and observes a reward from $D$ to update its policy.
    
    Note that there are two policies in COPT: the target policy that we aim to learn and the behavior policy used for the reasoning of scenarios.
    The behavior policy is pre-trained and then froze during adversarial learning because it aims to maximize the likelihood of a fixed set of observed responses.
    Introducing the behavior policy makes COPT an off-policy approach because the counterfactual response, from which the target policy learns, is not entirely based on the target policy itself.
    
    The goal of the generator is to minimize the negative expected reward: $J_G(\bm{\theta}) = - \mathbb{E}_{\bm{\hat{Y}}_{1:j} \sim G} D(\hat{Y}_{j}|\bm{X}, \bm{\hat{Y}}_{1:j-1})$, where $\bm{\theta}$ is the parameters of $\pi$.
    The gradient of $\bm{\theta}$ can be derived by the likelihood ratio trick~\cite{williams1992simple}:
    \begin{align} \label{gupdate}
    \bigtriangledown J_G(\bm{\theta}) = &- \mathbb{E}_{\bm{\hat{Y}}_{1:j} \sim G} D(\hat{Y}_{j}|\bm{X}, \bm{\hat{Y}}_{1:j-1}) \nonumber \\
    & \cdot \bigtriangledown \log G_{\pi}(\hat{Y}_j|\bm{X}, \bm{\hat{Y}}_{1:j-1}),
    \end{align}
    where $G_{\pi}(\hat{Y}_j|\bm{X}, \bm{\hat{Y}}_{1:j-1})$ is the probability of generating $\hat{Y}_{j}$ with the policy $\pi$ given $\bm{X}$ and $\bm{\hat{Y}}_{1:j-1}$.
    
    \begin{algorithm}[!t]
        \caption{\label{algo}Counterfactual Off-Policy Training} 
        \label{alg:Framwork} 
        \begin{algorithmic}[1]
            \STATE Pre-train $\pi$ and $\mu$ with MLE loss;
            \STATE Pre-train $D$ on positive instances sampled from observed responses, and negative instances generated by pre-trained $\pi$;
            \FOR{epoch in number of epochs}
            \FOR{$g$ in g-steps}
            \STATE Infer $\bm{u}$ from an observed response;
            \STATE Generate a counterfactual response in $\bm{u}$;
            \STATE Optimize $\bm{\theta}$ according to Equation~\ref{gupdate};
            \ENDFOR
            \FOR{$d$ in d-steps}
            \STATE Sample positive instances from observed responses;
            \STATE Sample negative instances from responses generated by $\pi$;
            \STATE Update $\bm{\phi}$ according to Equation \ref{d_loss};
            \ENDFOR
            \ENDFOR
        \end{algorithmic}
    \end{algorithm}
    
    The discriminator distinguishes between observed responses and model-generated responses.
    This is achieved by minimizing the following loss:
    \begin{align} \label{d_loss}
    J_D(\bm{\phi}) = &- \mathbb{E}_{\bm{Y}_{1:j} \sim \bm{S}}\log D(Y_{j}|\bm{X}, \bm{Y}_{1:j-1}) \\ \nonumber
    &- \mathbb{E}_{\bm{\hat{Y}}_{1:j} \sim G}\log (1-D(\hat{Y}_{j}|\bm{X}, \bm{\hat{Y}}_{1:j-1})),
    \end{align}
    where $\bm{\phi}$ is the parameters of $D$.
    As a positive instance, $\bm{Y}_{1:j}$ is a prefix randomly sampled from observed response set $\bm{S}$.
    A negative instance $\bm{\hat{Y}}_{1:j}$ for training $D$ is a prefix of a standard response, rather than a counterfactual response.
    This is because the latter is of higher quality than the former (as shown in Sec.~\ref{sec_analysis}).
    
    \paragraph{Pre-training} Initialized with different parameters, $\pi$ and $\mu$ are pre-trained on the training set with MLE loss.
    The pre-training of $D$ depends on the specific model that COPT applied to.
    For example, REGS pre-trains $D$ on the prefix of a response.
    In contrast, the discriminator of StepGAN~\cite{tuan2019improving} is randomly initialized during the adversarial learning process.
    The overall algorithm of COPT is summarized as Algorithm~\ref{algo}.

    \section{Experiments}
    \subsection{Data}
    The experiments are conducted on the DailyDialog dataset~\cite{li2017dailydialog}.\footnote{{http://yanran.li/dailydialog}}
    It is a multi-turn dialogue dataset and covers various topics of daily life.
    The dataset has already been divided into training, validation, and test sets, as shown in Table~\ref{dailydialogue}.
    Given a dialogue that consists of $K$ utterances, we divide it into $K$-1 instances.
    Each instance has at most three continuous utterances.
    The last utterance is the response, and the previous utterances are concatenated as the dialogue history.
  
      \begin{table}
        \centering
        \begin{tabular}{rr}
            \toprule
            Training Dialogues & 11,118 \\
            Validation Dialogues & 1,000\\
            Test Dialogues & 1,000\\
            Average Tokens Per Dialogue & 114.7\\
            Average Tokens Per Utterance & 14.6\\
            \bottomrule
        \end{tabular}
        \caption{\label{dailydialogue} Statistics of the DailyDialog dataset.
        }
    \end{table}

    \subsection{Baselines}
    We compare COPT with the following dialogue generation models:
    \begin{itemize}
        \item HRED~\cite{serban2016hred}: The hierarchical recurrent encoder-decoder.
        An implementation by~\newcite{park-etal-2018-hierarchical} is available\footnote{https://github.com/ctr4si/A-Hierarchical-Latent-Structure-for-Variational-Conversation-Modeling}.
        \item REGS~\cite{li2017adversarial}: Reward for every generation step. Its discriminator is trained on partially generated responses to provide a reward for each generation step.
        \item DPGAN~\cite{xu2018diversity}: The diversity-promoting GAN introduces a language model based discriminator to encourage the generation of informative responses.\footnote{{https://github.com/lancopku/DPGAN}}
        \item StepGAN~\cite{tuan2019improving}: The stepwise GAN trains the discriminator by maximizing the average of state-action values of observed responses.
        During the adversarial learning process, the discriminator assigns scores for every generation step in the same way as REGS. 
        
    \end{itemize}
    Distinct from previous approaches, COPT casts a dialogue generation model as an SCM and trains it on counterfactual responses.
    It is model-agnostic and can be applied to any adversarial learning-based dialogue generation model, such as REGS, DPGAN, and StepGAN.

    \begin{table}
    \centering
        \begin{tabular}{lr}
            \toprule
            Model & Time (s/epoch)\\
            \midrule
            HRED & 84 \\
            DPGAN & 608\\
            REGS & 912\\
            REGS+COPT & 1,215\\
            StepGAN & 951\\
            StepGAN+COPT & 1,244\\
            \bottomrule
        \end{tabular}
        \caption{\label{time} The average training time (in seconds per epoch) on a single GPU.
        }
    \end{table}
    
    \begin{table*} \small
        \centering
        \begin{tabular}{l|c|c|c|c|c|c}
            \toprule
            Model & Dist-1 & Dist-2 & BLEU-1 & BLEU-2 & BLEU-3 & BLEU-4\\
            \midrule
            HRED~\cite{serban2016hred} & 0.011 & 0.045 & 33.01 & 4.54 & 1.05 & 0.29\\
            DPGAN~\cite{xu2018diversity}& 0.002 & 0.008  &31.61 &3.74 & 0.43 &0.06 \\ \midrule
            REGS~\cite{li2017adversarial} & 0.021  & 0.097 &38.39 & 6.85 & 2.00 & 0.68 \\
            REGS+COPT & $\bm{0.026}$ & 0.116 & 39.80 & 7.67 & 2.34 & 0.77 \\ \midrule
            StepGAN~\cite{tuan2019improving} &0.018 & 0.097 &34.76 & 6.51 & 1.97 & 0.60\\
            StepGAN+COPT & $\bm{0.026}$ & $\bm{0.117}$ & $\bm{43.83}$ & $\bm{8.43}$ & $\bm{2.66}$ & $\bm{0.95}$ \\
            \bottomrule
        \end{tabular}
        \caption{\label{auto_eval} Automatic evaluation results of distinct-1 (Dist-1), distinct-2 (Dist-2), and BLEU scores.
        }
    \end{table*}
        
    \begin{table*} [!ht]
        \centering
        \begin{tabular}{ll}
            \toprule
            Dialogue History & A: This is our counter-offer. \\
            & B: We are sincere to make business with you, but the difference between your \\
            & counter-offer and our price is too great.\\
            \midrule
            HRED & A: We can do that, but you can take a look at this one.\\
            DPGAN & A: Think that the will be, but he said that they had a think we have a good job\\
            & in a. Me, I am sure that I can be able to have a look at the first course. \\
            REGS & A: If you are willing to withdraw money, we can offer you a private vehicle.\\
            REGS+COPT & A: If you are willing to pay a commission fee, we can offer you \$5 \% discount.\\
            StepGAN & A: What do you mean by that?\\
            StepGAN+COPT & A: In that case, we can offer you a discount.\\
            \bottomrule
        \end{tabular}
        \caption{\label{cs} An example of generated responses given dialogue history between person A and B.
        }
    \end{table*}
   
    \subsection{Training Details}
    We implement REGS, StepGAN, and their variants with COPT using OpenNMT~\cite{klein2017opennmt}, an open-source framework for building sequence to sequence models.
    We manually tune the parameters according to the perplexity on the validation set.
    The vocabulary consists of the most frequent 10,000 words.
    Including more words (up to 17,438, the total number of DailyDialog vocabulary) observes no improvement but takes more time for training.
    We use 300 dimensional GloVe~\cite{pennington-etal-2014-glove} vectors to initialize word embeddings.
    Both the encoder and the decoder are a two-layer LSTM in $G$ and a single layer LSTM in $D$.
    The number of hidden units is 500.
    
    During the adversarial learning process,
    we use the ADAM algorithm to alternately optimize $G$ and $D$ for one batch and five batches.
    The batch size is 64.
    We have tested the learning rate from 1e-6 to 1e-3. 
    REGS+COPT and StepGAN+COPT achieve the best performance on 1e-5.
    The number of parameters for all the baselines is in a range of 21M to 26M.
    Equipping an adversarial learning baseline with COPT will introduce extra parameters with the same amount of the generator's parameters.
    Contributed by the behavior policy, the parameters are learned by pre-training, and COPT will not increase the number of trainable parameters in adversarial learning.
    Table~\ref{time} shows the average training time.
    COPT may increase the training time due to the posterior inference of scenarios.
    But it facilitates the exploration of the high-reward area of the potential response space and subsequently improves the quality of responses.
    
    \subsection{Evaluation Metrics}
    \paragraph{Automatic Evaluation}
    We evaluate the diversity and the relevance of generated responses using  {\it distinct}~\cite{li2016mmi} and {\it BLEU}~\cite{papineni-etal-2002-bleu}, respectively.
    The distinct-$k$ is the number of distinct $k$-grams normalized by the number of words of responses.
    Since BLEU might correlate weakly with human judgments of quality in the single-reference setting~\cite{liu-etal-2016-evaluate}, we use the multi-reference DailyDialog test set~\cite{gupta-etal-2019-investigating}, where each instance is augmented with four human-written diverse responses.\footnote{{https://github.com/prakharguptaz/multirefeval}}
    
    \paragraph{Human Evaluation}
    The human evaluation is conducted on 200 instances randomly sampled from the test set.
    We create a project on Amazon Mechanical Turk~\cite{buhrmester2016amazon} (AMT) and employ five AMT workers to give a preference between two responses generated by our approach and a baseline.\footnote{{https://requester.mturk.com/}}
    To maintain the quality of the evaluation, the task is visible to workers whose approve rate is greater than 95\%, and the number of approved is greater than 500.

    \begin{figure*}[!ht]
        \centering
        \includegraphics[width=420pt]{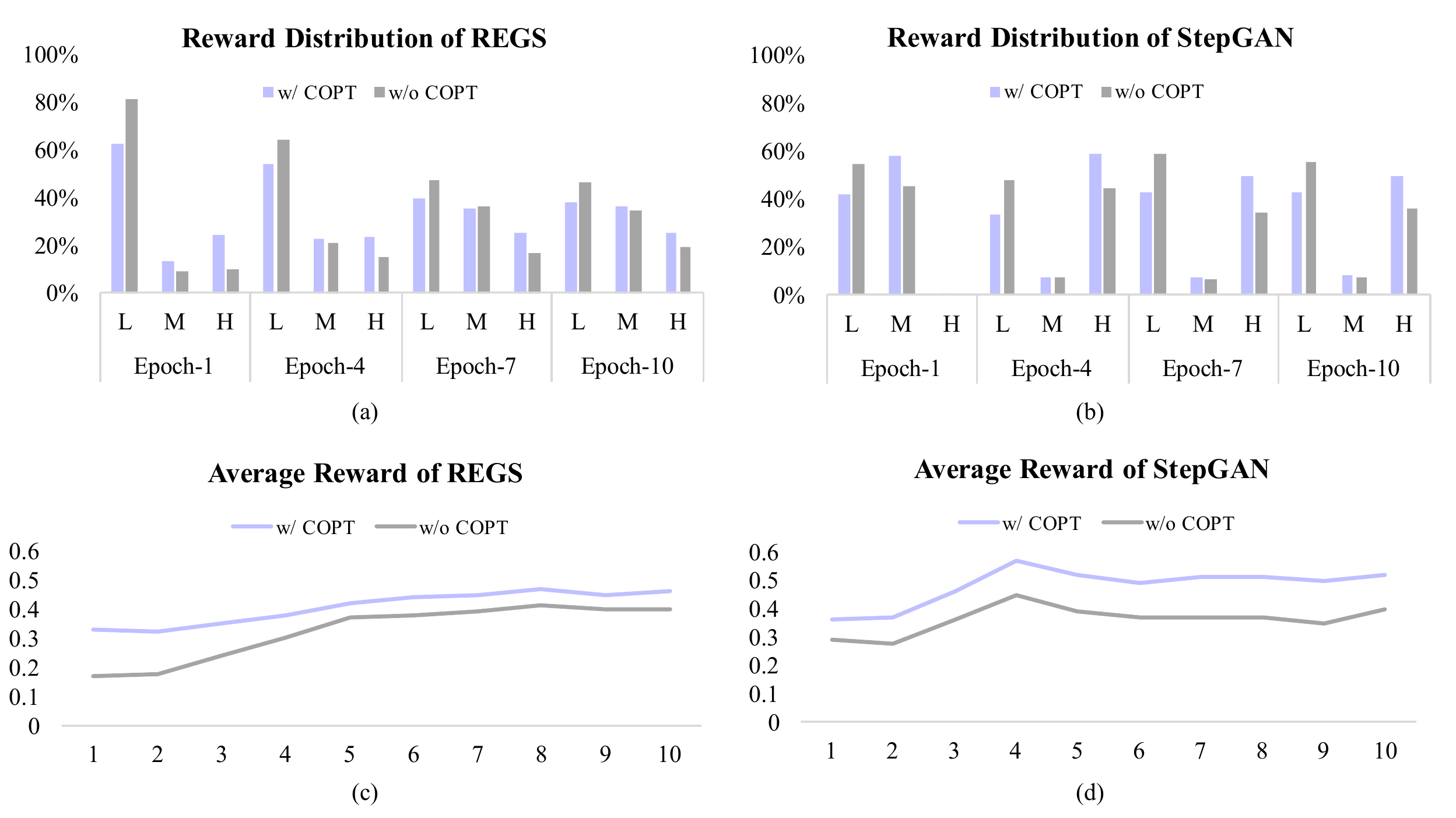}
        \caption{\label{analysis} 
            Reward distribution and the average reward of the counterfactual response (w/ COPT) and the standard response (w/o COPT).
            The y-axis in (a) and (b) is the percentage, and the x-axis corresponds to three reward intervals in different epochs, including: \textbf{L}ow reward interval [0.00, 0.33], \textbf{M}iddle reward interval (0.33, 0.66], and \textbf{H}igh reward interval (0.66, 1.00].
            The y-axis in (c) and (d) is the reward, and the x-axis corresponds to epochs.
        }
        
    \end{figure*}

    \subsection{Results} \label{sec_result}
    Table~\ref{auto_eval} shows the results of automatic evaluation.
    Both REGS and StepGAN outperform HRED in distinct-1 and distinct-2, indicating that adversarial learning is beneficial for improving the diversity of responses.
    There is no increase in DPGAN compared with HRED in our experiments.
    We believe this is because the scale of the DailyDialog dataset is not large enough for sufficiently training the language model based discriminator.
    For the same reason, COPT is not added to DPGAN.
    After introducing COPT, both distinct-1 and distinct-2 in REGS and StepGAN further increase, and the improvement is significant (t-test, $p$ \textless 0.01). 
    This suggests that COPT is model-agnostic to adversarial learning-based approaches and helps to promote the diversity.
    In terms of BLEU in Table~\ref{auto_eval},
    both REGS and StepGAN achieve higher BLEU scores with COPT, and the improvements of BLEU-1 and BLEU-2 are significant ($p$ \textless 0.05).
    This demonstrates the effectiveness of COPT in improving the relevance of responses.
    The less significant result of BLEU-3 and BLEU-4 is mainly due to the sparsity of tri-grams and four-grams, which are harder to be covered by references than uni-grams and bi-grams.
    
    The human evaluation results are shown in Table~\ref{human_eval}.
    Our approach is clearly preferred as it has more winning instances than losing instances ($p<$0.01).
    The results indicate that COPT helps improve the quality of responses.
    Following~\newcite{zhou2018commonsense} and \newcite{ke-etal-2018-generating}, we measure the agreement of annotators using inter-rater consistency.
    The percentage of instances that at least three annotators have the same preference (3/5 agreement) is 84.18\%.
    The percentage for 4/5 agreement is 46.89\%.
    
    \begin{table} \small
        \centering
        \begin{tabular}{l|ccc}
            \toprule
             & Win & Tie & Lose\\
            \midrule
            REGS+COPT vs. HRED & 63.15 & 13.20 & 23.65 \\
            REGS+COPT vs. REGS & 32.28 & 43.81 & 23.92 \\
            StepGAN+COPT vs. HRED & 65.95 & 15.38 & 18.67 \\
            StepGAN+COPT vs. StepGAN & 45.69 & 20.91 & 33.40 \\
            \bottomrule
        \end{tabular}
        \caption{\label{human_eval} Wins, losses, and ties (in \%) of our approach against baselines based on the human evaluation.
        }
    \end{table}

    \subsection{Case Study}
    Table~\ref{cs} shows an example of responses generated by baselines and our approach.
    The response of DPGAN sometimes is not fluent and can be very long.
    We believe this is also because the scale of the DailyDialog dataset is not enough for the language model discriminator.
    The response of HRED is not as informative as that of our approach. Its first part is generic, and what the pronoun ``that'' refers to is not clear.
    The response of StepGAN is not informative enough as well.
    In contrast, the response of REGS is quite informative, but its content is not entirely relevant to the dialogue history.
    After introducing COPT, the responses of REGS+COPT and StepGAN+COPT propose offering a discount to address Person B's concern of the price, which is both informative and relevant.
    
    \subsection{Analysis} \label{sec_analysis}
    To further analyze COPT's effectiveness in exploring the high-reward area of the potential response space during the training process, we compare the reward of a counterfactual response and a standard response on the same 10,000 randomly sampled training instances.
    However, the comparison between the two types of responses could be biased if their rewards are computed by different discriminators.
    Besides, the quality of responses is determined not only by the way they generated (with or without COPT) but also by the generator.
    To focus on the analysis of COPT and eliminate the bias between generators and discriminators, we generate and evaluate the two types of responses using an identical generator and its corresponding discriminator.
    Here, we use REGS+COPT and StepGAN+COPT as testbeds because they could generate both the two types of responses.
    
    Figure~\ref{analysis} shows the distribution of rewards and the average reward.
    The percentage of counterfactual responses in the high reward interval (0.66, 1.00] is higher than that of standard responses.
    Meanwhile, counterfactual responses generated with COPT achieve a higher average than standard responses.
    The results demonstrate the effectiveness of the counterfactual response in exploring the high-reward area of the potential response space during the training process.
    Note that the distribution and the average between different epochs are not comparable due to the update of the discriminator as the training processes.
    
    \section{Conclusion}
    We propose a model-agnostic approach, COPT, that can be applied to any adversarial learning-based dialogue generation models. 
    In contrast to existing approaches, it learns on counterfactual responses inferred from the structural causal model, taking advantage of observed responses.
    This helps the model to explore the high-reward area of the potential response space.
    Experiments show that the COPT significantly improves the quality of the generated responses, which demonstrates the effectiveness of this approach.
    
    \section*{Acknowledgments}
    The authors would like to thank all the anonymous reviewers for their insightful comments.
    This paper is supported by the National Natural Science Foundation of China under Grant No. 62076081, No. 61772153, and No. 61936010.
    The author from UCSB is not supported by any of the projects above.
    \bibliography{anthology,emnlp2020}
    \bibliographystyle{acl_natbib}
    
    \appendix

\end{document}